\def\papertitle{Genetic Algorithms and the Traveling Salesman Problem a historical Review}
\def\paperauthorA{Jan Scholz}

\documentclass[twoside,a4paper]{article}
\usepackage{minfsem}
\usepackage{placeins}
\usepackage{amsmath,amssymb,amsfonts,amsthm}
\usepackage{euscript}
\usepackage[utf8]{inputenc}
\usepackage[T1]{fontenc}
\usepackage{ifpdf}
\usepackage{booktabs} 
\usepackage[para,online,flushleft]{threeparttable}
\usepackage{enumitem}

\usepackage[english]{babel}
\usepackage{caption}
\usepackage{subfig, color}

\usepackage{blindtext}
\usepackage{lipsum}

\ninept

\usepackage{times}

\newif\ifpdf
\ifx\pdfoutput\relax
\else
   \ifcase\pdfoutput
      \pdffalse
   \else
      \pdftrue
\fi

\ifpdf 
\PassOptionsToPackage{hyphens}{url}
  \usepackage[pdftex,
    pdftitle={\papertitle},
    pdfauthor={\paperauthorA},
    colorlinks=false, 
    bookmarksnumbered, 
    pdfstartview=XYZ 
  ]{hyperref}
  \usepackage[pdftex]{graphicx}
  \usepackage[figure,table]{hypcap}
\else 
  \usepackage[dvips]{epsfig,graphicx}
  \usepackage[dvips,
    colorlinks=false, 
    bookmarksnumbered, 
    pdfstartview=XYZ 
  ]{hyperref}
  \usepackage[figure,table]{hypcap}
\fi

\title{\papertitle}

\affiliation{
\paperauthorA, }
{\href{http://www.haw-hamburg.de/ti-i}{Hamburg University of Applied Sciences,
    Dept. Computer Science,} \\ Berliner Tor 7\\ 20099 Hamburg, Germany\\
{\ttfamily \href{mailto:jan.scholz2@haw-hamburg.de}{jan.scholz2@haw-hamburg.de}}
}

\begin{document}
\ifpdf 
  \DeclareGraphicsExtensions{.png,.jpg,.pdf}
\else  
  \DeclareGraphicsExtensions{.eps}
\fi

\maketitle

\begin{abstract}
In this paper a highly abstracted view on the historical development of Genetic Algorithms for the Traveling Salesman Problem is given. In a meta-data analysis three phases in the development can be distinguished. First exponential growth in interest till 1996 can be observed, growth stays linear till 2011 and after that publications deteriorate. These three phases are examined and the major milestones are presented. Lastly an outlook to future work in this field is infered.
\end{abstract}

\section{Introduction}
\label{sec:Introduction}






The Traveling Salesman Problem (TSP) is a long known problem habituated in the NP-Hard complexity space. The problem has been excessively studied\cite{karp1972reducibility}\cite{applegate2006}\cite{miller1960integer}\cite{dantzig1954solution}\cite{fischetti1997branch}\cite{lin1973effective} and a vast array of methods have been introduced to either find the optimal tour or a good less time consuming approximation. This paper will concentrate on the second path of meta-heuristics and specifically on genetic algorithms (GA) and the historical association with the TSP.\\
GA's have been around since 1957\cite{fraser1957simulation}, starting with simulations for biological evolution. GA's are used for optimization problems with large search spaces. The TSP as an optimization problem therefore fits the usage and an application of GA's to the TSP was conceivable.\\
In 1975 Holland \cite{holland1975adaptation} laid the foundation for the success and the resulting interest in GA's. With his \textit{fundamental theorem of genetic algorithms} he proclaimed the efficiency of GA's for optimization problems. A generic GA starts with the generation of a population of several different tours. Two tours (parents) are combined with crossover\footnote{crossover: operation in which a new chromosome is created from two parent chromosomes. A chromosome is an encoded solution of the problem.}, the conceived child may experience mutation\footnote{mutation: alteration of the chromosome} and is then given a fitness-score, in case of the TSP the tour length. On basis of the fitness-score a part of the population is discarded and the rest is used to create new offspring. GA's\footnote{All further mentions of GA's are all to be seen in the context of the TSP unless stated otherwise.} are tightly interwoven with local search\footnote{local search algorithms are characterized by iteratively changing solutions to find better solutions }, together they perform better than local search alone, this is because local search loses itself in local optima. Other meta-heuristics like simulated annealing serve the same purpose as GA's but GA's have the inherently good capability to merge tours and develop a better tour from good sub tours.\\
In 1985 Brady\cite{brady1985optimization} introduced the first GA for the TSP. The implementations for GA's changed over time, from generic to more specific to the TSP and it's subcategories of problems. Over the years several encodings\footnote{encoding: The representation of the chromosome (genes) in form of a tour}, crossover- and mutation-methods were invented which further specialized GA's for the TSP.\\
It will be shown that GA's in context of the TSP have been researched extensively in the past but the interest for this field seems to vanish. The reasons for this fall in appeal of GA's will be examined. Further will be shown which methods prevailed over the years and how the future of GA's for the TSP might develop.\\




\section{Related Work}
\label{sec:RelatedWork}
In the research for this paper there were no publications that fit exactly on the topic of this work. But there are a few noteworthy publications that review GA's and give a comprehensive overview of the field for the given time. These works are useful to get an abstracted view into the historic development of GA's for different time periods and also occasionally show computational results for a variety of GA's.\\
The earliest is from 1988 by Fogel\cite{fogel1988evolutionary} in which he presents the early efforts of GA's. He emphasizes the importance of crossover for GA's, without crossover the algorithm wouldn't be more than a random search. But also that evolutionary stagnation\footnote{evolutionary stagnation is exchangeable with genetic depletion and leads to premature convergence} is inevitable as long as inheritance in the form of crossover is present. Fogel warns of over-ambitious mutation operators that could destroy the link between offspring and parent. The most important observation was that then current encodings weren't comparable to chromosomes of organisms and thus crossover methods that try to replicate biological crossover are unsuitable.\\
In Potvin's\cite{potvin1996genetic} (1996) paper several crossover methods are presented and categorized into 3 groups: Relative order, Position and Edge. The main points discovered by Potvin were, edge-preserving crossover outperforms other crossover operators, also it has been noted that local hill-climbing seemed to be crucial for good performance. The separation of the population into sub-populations that sparingly mix is useful to prevent premature convergence. Also larger populations correspond to better solutions. Potvin prominently features the matrix-based encoding in his paper and mentions that parallel GA's will greatly improve solutions in the future.\\
In 2001 Merz \& Freisleben\cite{merz2001memetic} give a detailed summary of the most effective GA's. They particularly note edge assembly crossover as highly effective. They identify GA's as very well suited for the search space landscape of the TSP, this is because local optima are concentrated in a cluster for randomly generated TSP's. Also the authors claim that besides the branch and cut approach from Applegate et al.\cite{applegate2006} and GA's there were no known heuristics that can solve larger TSP instances (>3000 nodes).\\
 
\section{Material and Methods}
\label{material_methods}

\begin{table}[]
    \centering
    \caption{Sources}
    \label{tab:sources}
    \begin{tabular}{@{}lll@{}}
    \toprule
    Source  & Type & amount \\ \midrule
    dblp\cite{dblp} & publications/year & -   \vspace{2mm} \\
    acm\cite{acmdl} & papers & $\sim$10  \vspace{2mm} \\
    semanticscholar\cite{semanticscholar} &
    \begin{tabular}[c]{@{}l@{}}meta-data\\ papers \end{tabular}  &
    \begin{tabular}[c]{@{}l@{}}26940\\ $\sim$20 \end{tabular}  \vspace{2mm} \\
    BGP\cite{gabib}  & meta-data & 13121  \\ \bottomrule
    \end{tabular}
\end{table}

The used data is constituted of the content of papers, books and articles and also the meta-data of publications for GA's. The integral papers for genetic Algorithms and the TSP are read and the key insights are discovered. With the meta-data several measures are looked into to understand the development of genetic algorithms. The used metrics are publication-frequency for papers regarding TSP and GA's and mentions of specific terms in the abstracts from those publications.\\
The data from semanticscholar was acquired with several search queries on the open API\footnote{example of a searchquery for the term ``genetic 1999'' looking at the results 1001 to 2000 \href{https://www.semanticscholar.org/api/1/partner/dblp/search?q=genetic\%7C1999\&s=1000\&p=2}{here}} the returned format is json and is easily computed for further inquiry. The search terms are formed by a list of keywords\footnote{keywords are: genetic, ga, evolutionary, natural, crossover, mutation, traveling, travelling, salesman, tsp, salesperson, combinatorial optimization} in combination with all years from 1970 to 2017. With these queries 26940 unique entries were gained.\\
The dataset from the Bibliography on Genetic Programming had an artifact, a strong decline from 1999 to 2000 in GA publications could be seen. Before the existence of the artifact was known the initial notion was to look for the cause of this sudden decline, this question can now be considered resolved. The reason was an inflated size of publications till 1999, all conference papers of the GECCO were automatically added\cite{langdonpersonal} before 2000. So a large chunk of publication weren't strictly related to genetic programming. The database was filtered and every publication without the keyword ``genetic programming'' was discarded, about 300 publications in 1999 and 680 overall were filtered. The resulting graph can be seen in Figure \ref{fig:gapub}, it may not represent the actual circumstances but it's a closer guess than the previous data.\\
The database from semanticscholar forms a metric called \textit{influential citations} this metric has the purpose to indicate works which are meaningful to other publications. This metric was conceived by Valenzuela et al.\cite{valenzuela2015identifying} and uses 12 different features such as PageRank, Author-overlap and similarity of abstracts to be computed. In Figure \ref{fig:gapoppubs} 64 publications are shown with the corresponding metric. This set of publications is a subset of 179 publications which are found with a title that refers to the TSP and GA's, they additionally have a greater than 0 value for \textit{influential citations}. It can be seen that 7 publications do stand out with a score more than 11 and up to 36. These 7 publications can be considered most important and were therefore exhaustively studied.

\subsection{Limitations}
The meta-data from semanticscholar is drawn from a database of over 40Mio entries, which draws from several other big databases, to the knowledge of the author it is the most comprehensive archive of it's kind. The database is vast but incomplete, this applies especially to chapters in books, journal articles and lesser known publications. In the research for this topic out of 68 publications 10 were not findable\footnote{untraceable publications by type: 3 articles in journals, 3 books, 2 chapters in books, 1 conference paper} in the database of semanticscholar. It has been positively noted that all essential publications were traceable. \\
For all sources applies that publications from recent years are more heavily impartial because the time for them to be discovered is less than for older publications. This can be seen in Figure \ref{fig:gapub}, in the year 2017 have been supposedly lesser papers published than in 2016. This will likely be corrected to a higher number in the following months.\\
The analysis of meta-data may only act as an indicator for development because some publications that are strictly to GA's and the TSP may not give a descriptive title that contemplates this. Also some meta-data entries may not have the abstract for the publication stored, this should be a bigger problem for older publications which are not digital but scanned and may have not been interpreted with OCR.\\
The comparability in quality and computation time of different GA implementations is very delicate because there are innumerous side-conditions that are linked to both. To list a few: hardware, quality of implementation, Programming Language, parallel or sequential design and TSP instances with equal nodes can differ enormously in difficulty. To account for this, studies where various implementations are tested by the author are preferable to asses this question.

\section{Evaluation}
\label{sec:evaluation}

\begin{figure*}[ht]
    \includegraphics[width=\textwidth]{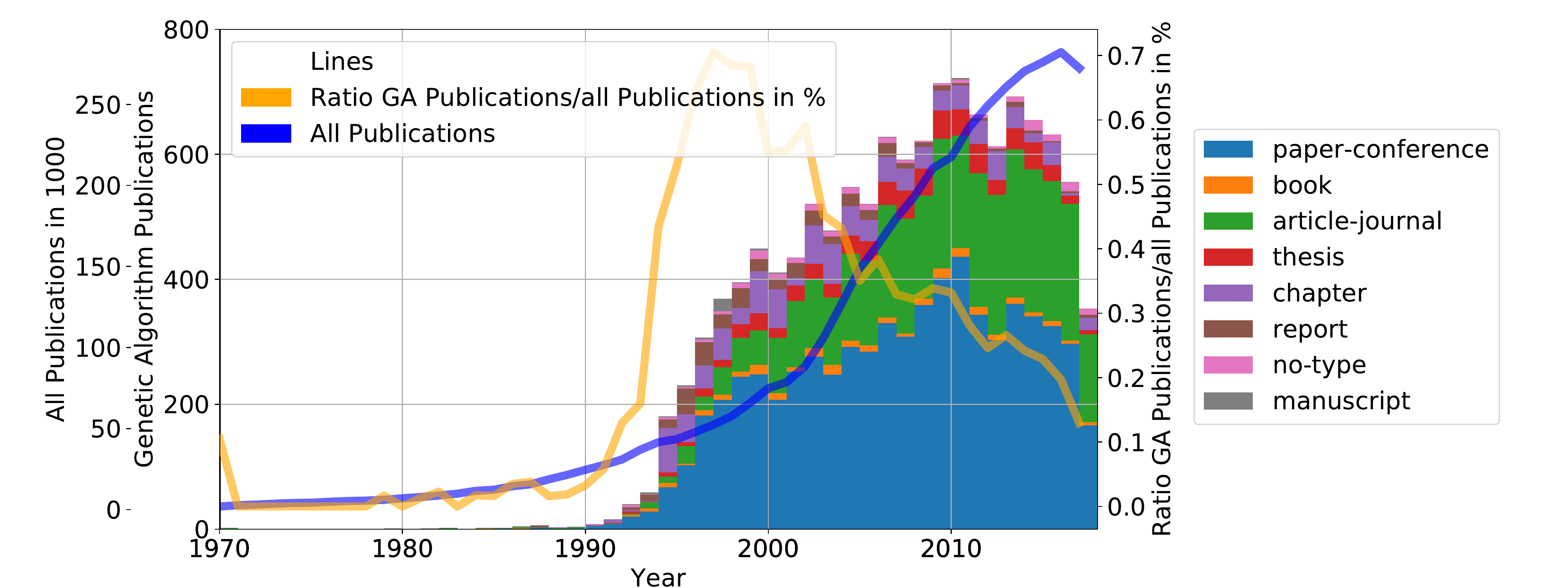}
    \caption{Publications for GA in comparison to publications in computer-science in general}
    \label{fig:gapub}
\end{figure*}

\begin{figure*}[ht]
    \includegraphics[width=\textwidth]{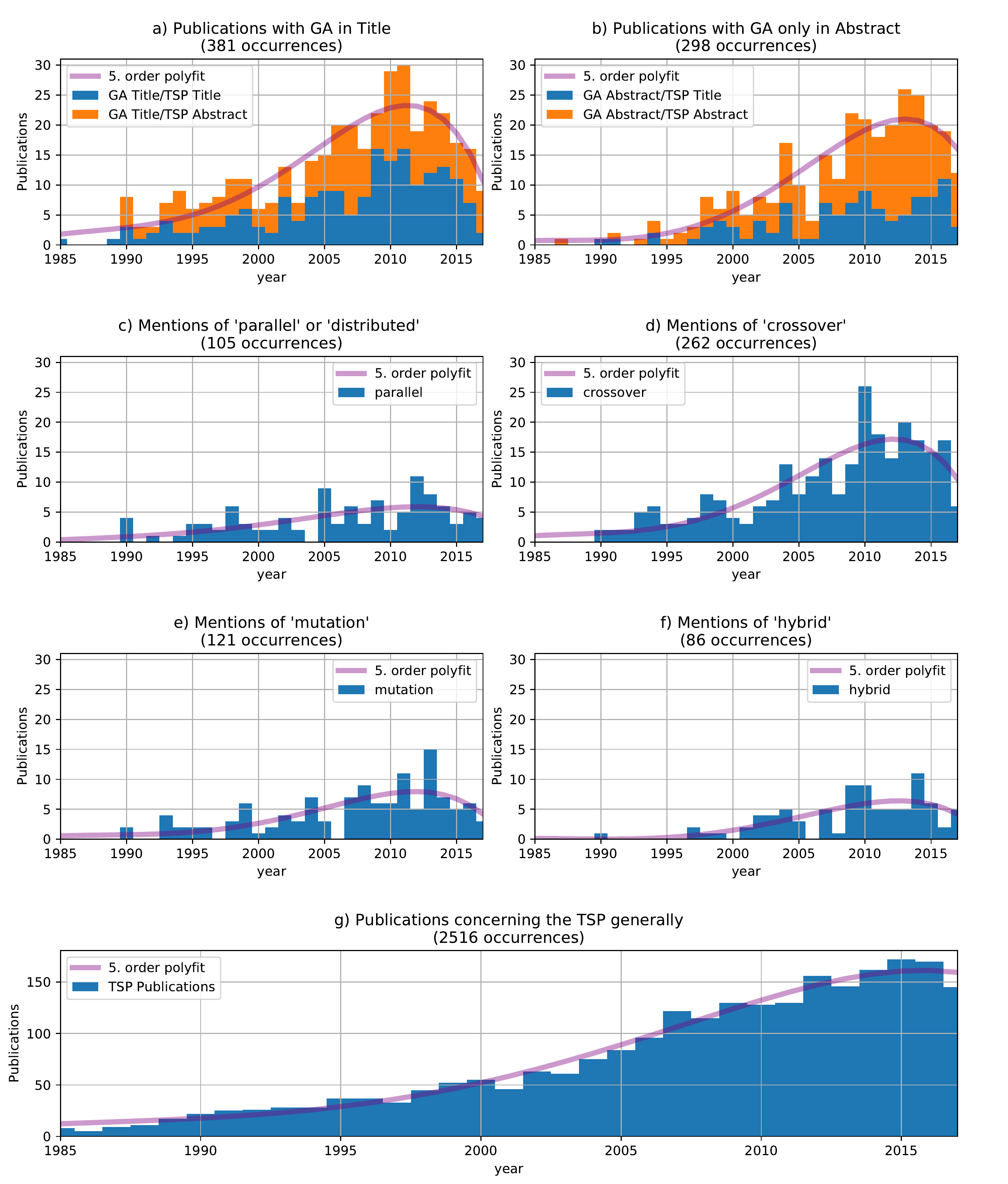}
    \caption{Publication metrics concerning TSP and GA}
    \label{fig:gatsppub}
\end{figure*}

The development of GA's will be split into 3 epochs which are separated by distinct events. The first phase ends with the beginning decline of the ratio GA (all not only TSP) publications / computer-science publications (see Figure \ref{fig:gapub} orange line with maximum at 1996) at this time the growth changes from exponential to linear. The third phase starts in 2011 when absolute publications for GA's start to decline (see Figure \ref{fig:gatsppub}a).

\begin{table*}[htb]
    \centering
    \caption{Timeline with Achievements}
    \label{tab:achieve}
    \begin{threeparttable}
    \begin{tabular}{@{}lll@{}}
    \toprule
    Year  & Author & Achievement/Discovery \\ \midrule
    1985  & Brady\cite{brady1985optimization}  &  first use of GA for the tsp, 2-opt used for optimization \\
    1985  & Grefenstette\cite{grefenstette1985genetic}  & heuristic crossover \\
    1985  & Goldberg \& Lingle\cite{Goldberg1985AllelesLociandTT}  & Partially-mapped crossover \\
    1987  & Oliver et al.\cite{oliver1987study}  & Order crossover, Cycle crossover \\
    1988  & Mühlenbein et al.\cite{muhlenbein1988evolution} & \begin{tabular}[c]{@{}l@{}}first parallel algorithm with super linear\tnote{1} \\ performance also donator/receiver crossover \end{tabular}  \\
    1988  & Fogel\cite{fogel1988evolutionary} & biological crossover unsuitable for tsp (path encoding)\tnote{2} \\
    1990  & Braun\cite{braun1990solving} & insular genetic algorithm\tnote{3} \\
    1991  & Syswerda\cite{syswerda1991scheduling} & Position based crossover \\
    1990  & Johnson\cite{johnson1990local} & \begin{tabular}[c]{@{}l@{}}iterated lkh, with no crossover computes \\ larger instances in reasonable time\tnote{4} \end{tabular}\\
    1991  & Whitley et al.\cite{whitley1991traveling} & Edge recombination crossover \\
    1992  & Homaifar\cite{homaifar1992schema} & Matrix crossover \\
    1994  & Srinivas \& Patnaik\cite{Srinivas1994AdaptivePO} & adaptive probabilities for crossover and mutation \\
    \midrule
    1996  & Merz\cite{freisleben1996genetic} & DPX crossover \\
    1997  & Nagata\cite{nagata1997edge} & Edge assembly crossover \\
    2002  & Nguyen\cite{nguyen2002greedy} & GSX based GENITOR-type GA\tnote{5}\\
    2006  & Snyder\cite{snyder2006random} & random-key encoding  \\
    2007  & Ray\cite{ray2007genetic} & modified order crossover, nearest fragment operator \\ \midrule
    2011  & Albayrak\cite{albayrak2011development} & Greedy Sub Tour Mutation \\
    2014  & Jäger et al.\cite{jager2014backbone} & pseudo-backbone edges \\
    \bottomrule
    \end{tabular}
    \begin{tablenotes}
        \scriptsize{\item[1] super-linearity is achieved with growing entities by a processes, this observation is not only due to the parallel nature of the algorithm but also because the population is larger and therefor more diverse} \\
        \scriptsize{\item[2] path encoding doesn't reflect the characteristics of chromosomes, therefore biological crossover is not applicable}\\
        \scriptsize{\item[3] multiple populations that only mix, to refresh the gene pool, when degradation occurs}\\
        \scriptsize{\item[4] random 10$^4$ city problem within 0.8\% lower-bound in ~75h}\\
        \scriptsize{\item[5] Nguyen found several optimal tours for Problems <10$^4$ and a solution for the World TSP (1.9Mio Nodes) within a 0,05\% error margin to the lower bound found by Concord}
    \end{tablenotes}
\end{threeparttable}
\end{table*}

\begin{figure}[htbp]
    \begin{minipage}{0.45\textwidth}
        \centering
        \includegraphics[width=\textwidth]{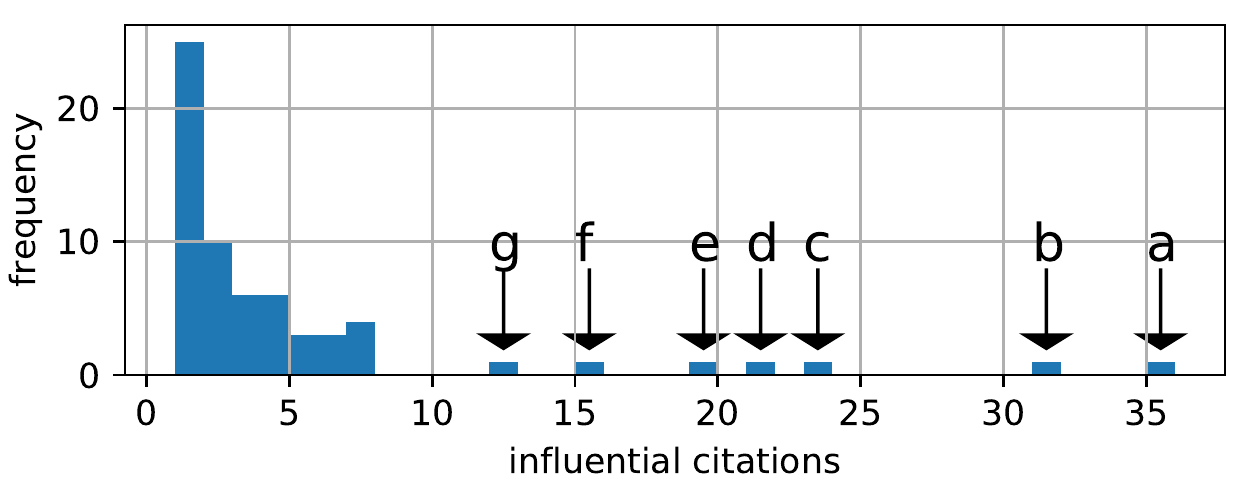}
    \end{minipage}
    \hfill
    \begin{minipage}{0.39\textwidth}
        \begin{enumerate}[label=\alph*)]
            \item Larrañaga et al. (1999) \cite{larranaga1999genetic}
            \item Freisleben \& Merz (1996) \cite{freisleben1996genetic}
            \item Nagata \& Kobayashi (1997) \cite{nagata1997edge}
            \item Grefenstette et al. (1985) \cite{grefenstette1985genetic}
            \item Snyder et al. (2006) \cite{snyder2006random}
            \item Albayrak \& Allahverdi (2011) \cite{albayrak2011development}
            \item Ray et al. (2007) \cite{ray2007genetic}
        \end{enumerate}
    \end{minipage}
    \caption{Frequency of influential citations for Publication regarding GA and TSP}
    \label{fig:gapoppubs}
\end{figure}

\vspace{-5pt}
\subsection{Inception (1985 - 1995)}
As mentioned the first GA was introduced by Brady (1985)\cite{brady1985optimization} this implementation already used local search to optimize tours and enhance the mutation process. Pure genetic algorithms were already sorted out in the beginning as a viable idea to efficiently solve TSP instances\cite{Grefenstette87incorporatingproblem}, the incorporation of local search and domain specific knowledge was imperative. In 1993 Koza\cite{koza1992genetic} published his book {\it Genetic Programming}, this book is the most significant publication for GA's in general (1275 influential citations). Shortly after the publication the technology took off and publications for GA's in general kept rising exponentially for a few years. This book inspired many scientists to engage with GA's also for solving the TSP.\\
The focus of the early years laid heavily on suitable crossover operators and encodings (see Table \ref{tab:achieve}). The publication in this period with the greatest impact (see Figure \ref{fig:gapoppubs}) was Grefenstette et al. (1985)\cite{grefenstette1985genetic} in which the heuristic crossover was presented, this crossover operator was the first to integrate problem specific knowledge. The entanglement between encoding and crossover is strong, a crossover operation has to be chosen in accordance to the encoding. Encodings for the TSP (i.e. path-encoding, matrix representation) did not allow for seamless crossover\cite{fogel1988evolutionary}. The most common encoding was the intuitive path-encoding in which every node was referenced by number and arranged in a sequence.\\
The advantage of genetic algorithms are that good qualities of the parents are passed on to the children with crossover. A pitfall of this is evolutionary stagnation\cite{fogel1988evolutionary} when progress converges to a local minimum. This could only be prevented with the omission of inheritance in the form of crossover. According to Holland \cite[P.110]{holland1992adaptation}, improvement without crossover is equal to a random-search, and follows an enumerative plan that is characterized by not considering outcomes of previous iterations in the current operation. Fogel (1988)\cite{fogel1988evolutionary} adds, however, that the limitation of the search space to the proximity of the parents still has an advantage. In 1990 Johnson\cite{johnson1990local} suggests the iterated Lin-Kernighan Algorithm (ILK) which abandoned crossover altogether and used the Lin-Kernighan Heuristic (LKH) for local optimization. This was a huge advancement for efficient computation of larger instances of the TSP. Beforehand it was not feasible to use GA's for instances larger than 1000 nodes (recalled by Potvin in 1996\cite{potvin1996genetic}). The ILK was then able to compute solutions within a Held-Karp lower bound of 0,8\% for instances up to 10$^4$ nodes, a computation for a 10$^4$ problem lasted 75h\footnote{computed with a VAX 8550 (22.2MHz, 1 core, 6.1 MIPS), this machine could then be considered a super-computer. To contrast this a newer model intel i5-2500k (3.5Ghz, 4 cores, 10262 MIPS)}. The concept of genetic algorithms is slightly subverted with the loss of crossover which is a fundamental quality of the evolutionary character of GA's and it depends on the used definition weather this implementation is still a GA.\\
Mutation is a key factor of the algorithm to search the complete event space of the problem. Holland sees mutation as not only necessary to provide all needed alleles\footnote{allele: gene-sequence, sub-section of a chromosome}\cite[P.110]{holland1992adaptation}, but also vital to reintroduce alleles that have been lost during the selection process but are included in the optimal solution. Changing the probability of mutation and crossover does help to converge in nontrivial multimodal functions according to Srinivas \& Patnaik\cite{srinivas1994genetic}. Early convergence to local minima is prohibited by the adaptive probabilities.
GA's are predestined for parallel computation\cite{muhlenbein1988evolution} because populations or even single entities can be split into different processes and eventually be merged. The first parallel algorithm was shown in 1988 by Mühlenbein et al.\cite{muhlenbein1988evolution}. Till 1995 Alba\cite{alba1999survey} cites 11 different proposed methods that are able to be processed parallel and some are worked out in great clusters of >100 machines. An implementation for FPGA's\footnote{Field Programmable Array: Integrated Circuit that can be programmed (with logic gates) compute massive parallel operations} by Graham \& Nelson\cite{graham1995hardware} went into another direction and parallelized the operations. The computational results showed that a similar sequential algorithm took 6 to 11 times longer to compute. But the results shouldn't be taken without caveat considering that the TSP instances didn't exceed 120 nodes, the tour quality wasn't stated and the computation time wasn't sensational\footnote{comparison Graham: 120 nodes, 296s Merz: kroA100.tsp, 11s, optimal; d198.tsp, 253s, optimal}. Conceivably there were no other noteworthy publications investigating this trend. Parallel computation is not the holy grail of performance for GA's the increase of population size doesn't boost\cite{jog1989effects} the performance linearly, the super linear behavior in \cite{muhlenbein1988evolution} can be accounted for the speed up in the seeding\footnote{seeding: the initialization of the first tour for an entity, often nearest-neighbor, 2-opt and other fast algorithms are used} process which takes a significant smaller share of the overall computation time for larger instances.\\


\vspace{-15pt}
\subsection{Improvement (1996 - 2010)}

The publications for GA's in general kept rising (see Figure \ref{fig:gatsppub}a) but the overall share of total publications in computer-science declined from 1995 onwards (see Figure \ref{fig:gapub}).
Since 1995 the keyword \textit{hybrid} is more often associated with GA's (see Figure \ref{fig:gatsppub}f) , it indicates the support of another algorithm (mostly local search) to enhance the performance. But literally the overwhelming majority of all preceding GA's could already be described as hybrid GA's. And it is again stated that pure genetic algorithms (non-Lamarckian) without local search can not achieve good results in a reasonable\cite[P.285]{johnson1997traveling}\cite{larranaga1999genetic} time. \\
As of notable Achievements there were three new crossover methods DPX\cite{freisleben1996genetic}, GX\cite{merz2001memetic} and EAX\cite{nagata1997edge} which were more sophisticated than previous crossover methods and could be categorized not as simple operators but algorithms in their own right. In 1999 Larrañaga \cite{larranaga1999genetic} wrote a review for encodings and crossover operators, this publication is highly respected and to this day a very good primer to the topic. The main take-away was that \textit{Order crossover}\cite{oliver1987study}, \textit{partial mapped crossover}\cite{syswerda1991scheduling} and \textit{edge recombination}\cite{whitley1991traveling} are superior operators\footnote{the  previously mentioned operators (DPX, EAX, GX) weren't included in this examination}. In 2007 Snyder et al.\cite{snyder2006random} introduced, the random-key encoding which enabled the development of new more organic crossover operators. \\
Nguyen\cite{nguyen2002greedy} computed with a GENITOR-type\cite{whitley1989genitor} GSX GA several previously unknown optimal solutions for instances up to 10$^4$ nodes\footnote{http://www.math.uwaterloo.ca/tsp/vlsi/summary.html}. In 2007 Ray\cite{ray2007genetic} introduced \textit{modified order crossover} and the \textit{nearest fragment operator}. In the same paper Ray gave a good comparison of his implementation and other high ranking implementations like LKH, Concord, ILK and Tabu search + LK. The central findings were that for all instances up to 13509 nodes his implementation had a lower error than all other algorithms with the same computation time. This further illustrates the competitiveness of GA's. In 2005 there was another approach from Vega et al.\cite{vega2005genetic} for a FPGA implementation which attracted some attention but did not perform reasonably well, this was due to the large amounts of memory accesses of the fitness function.


\vspace{-5pt}
\subsection{Maturity (2011 - present)}
After approximately 2011/2012 the peak of interest for GA's was exceeded and absolute publications started to decline (see Figure \ref{fig:gatsppub} a/b). This shall not be confused with a general decline in popularity for the TSP which may not even hit its peak by now (also Figure \ref{fig:gatsppub} g). Impactful publications in this period are sparsely sown hardly any achieve a substantial high number of citations. The decline of GA's in general coincides with the rise of machine learning (ML) which also solves optimization problems. However ML doesn't seem to be particularly good when directly applied to combinatorial optimization problems like the tsp\footnote{It's more common to use ML as Hyper-heuristic\cite{ozcan2010reinforcement}\cite{bai2005investigation} to infer good hyper parameters for heuristics.}.\\
The publication that got the most attention was Albayrak et al.\cite{albayrak2011development} with a new mutation operator Greedy Sub Tour Mutation. What strikes the eye is that mutation operators weren't as popular to begin with because the primary work between crossovers was usually done by local search optimization. Also worthwhile to mention is that none of the other inferior mutation operators he compared against was younger than 1992.\\
The former attempts to parallelize the GA with dedicated hardware were succeeded by few attempts which still didn't perform exceptionally. Implementations for GPU's forfeit customizability but are in one way ahead of FPGA implementations, they didn't suffer the problem of slow memory accesses due to the GPU architecture which supports fast access to huge chunks of memory. In 2010 Fujimoto et al.\cite{fujimoto2010highly} presents an implementation using CUDA for Nvidia cards, which unfortunately did not solve instances larger than 512 nodes because it was not possible for them to implement a local search algorithm in this environment. The best concept to this day seems to be from Kang et al.\cite{kang2016gpu} which did not use local search and concentrated the effort into a suitable crossover operator for parallel execution. However this algorithm was able to tackle larger instances and could approximate a solution for the Mona Lisa TSP\footnote{Mona Lisa TSP: 10$^5$ nodes\\http://www.math.uwaterloo.ca/tsp/data/ml/monalisa.html} with an error of only 0,065\% within about 70h.\\
In 2013 Nagata ties Xavier Clarist with his implementation\cite{nagata2013powerful} (EAX) for best solution of the 115475 USA city challenge. 


\section{Conclusion}
\label{sec:Conclusion}

The development of GA's has been overlooked from beginning to present. A shift from inherent GA operations to more and more problem specific knowledge exploitation is visible. The mayor milestones are highlighted and the timespan has been divided into 3 epochs. These periods show distinct features in the areas participation, performance gains and refinement of implementations. In the fist phase \textit{inception} a barrage of different ideas is introduced, of which few stand the test of time, implementations are simple and greater performance leaps are common. In phase two \textit{improvement} the proven good solutions lay the groundwork for more sophisticated algorithms that make small improvements or enhance only edge-cases of the problem. Ground braking new discoveries on which can be build upon are rare. In the current third phase \textit{maturity} it is immensely difficult to find better algorithms, opportunities for further optimization may lay in specialized hardware, but the interest and therefore scientific effort starts to decline. The reason for the decline in appeal of GA's seems to be that the field has been extensively studied and the investment of time and money doesn't seem to be worth it to find a just marginally better algorithm.\\
\subsection{Application of GA's today}
When heuristics are looked into computation-time and error are the most significant factors for comparison. If an optimal solution is necessary for a yet unknown TSP instance it is best to use an exact algorithm to know for certain that the solution is indeed optimal. Solutions provided by heuristics may be optimal but can not be verifiably optimal. If an application requires an ad-hoc solution to a TSP instance the lower bound could be approximated with a minimal 1-tree and a computational complexity of $O(m \alpha(m,n))$\footnote{Complexity of soft heap algorithm\cite{chazelle2000soft}: $O(m \alpha (m,n)$ $m$=weighted edges, $n$=vertices, $\alpha$=inverse Ackermann-function)} and after or parallel to that a GA could be employed till an acceptable solution is found. GA's have been employed for fast approximations and also to find optimal solutions. For fast approximation an IHK is a good answer and for very low error solutions a refined algorithm like Nagata's\cite{nagata2013powerful} is applicable.\\
\subsection{Future of GA's}
As earlier stated the development of GA's may have overstepped it's peak. The Future of GA's seems to hold decreasing innovation, similar to for example sorting algorithms that have come to almost optimal efficiency already in the years from 1945 (Merge-sort) to 1964 (Heap-sort). The later utilization of quantum computation may enable GA's to ascend but may also make other solutions more viable. Implementations for GPU are an entry point for possible high performance consumer applications, as almost every computer has a dedicated GPU. As of now there is no GPU implementation that can fully replicate the inner workings of a GA.\\

\FloatBarrier

\newpage
\bibliographystyle{IEEEbib}
\bibliography{ref}

\end{document}